# An ensemble-based online learning algorithm for streaming data

Tien Thanh Nguyen[1], Thi Thu Thuy Nguyen[1], Xuan Cuong Pham[2], Alan Wee-Chung Liew[1], James C. Bezdek[3]

[1]School of Information and Communication Technology, Griffith University, Australia

[2]Department of Computer Science, Water Resources University, Hanoi, Vietnam

[3]Computer Science and Information Technology Department, University of Melbourne, Australia

**Abstract**

In this study, we introduce an ensemble-based approach for online machine learning. The ensemble of base classifiers in our approach is obtained by learning Naïve Bayes classifiers on different training sets which are generated by projecting the original training set to lower dimensional space. We propose a mechanism to learn sequences of data using data chunks paradigm. The experiments conducted on a number of UCI datasets and one synthetic dataset demonstrate that the proposed approach performs significantly better than some well-known online learning algorithms.

1. Introduction

With rapid advances in storage and sensor technologies, large volumes of data in the form of data streams are being collected in many applications such as network traffic and stock market analysis. Streaming data has created problems for traditional offline machine learning systems. First, learning the entire volume of data at once to form the discriminative model is often not possible. Moreover, offline algorithms require re-training when new data are available, so they are not applicable for the situation where data is arriving continuously and the prediction model must be obtained before all data are available. Therefore, the online learning framework that deals with data streams has become increasing popular [Nguyen *et al.*, 2016a].

In this paper, we focus on supervised online learning, where the training data arrives sequentially. In online learning, learning only occurs, i.e. the prediction model is updated, when a label for the observation is made available. Otherwise, the algorithm is performing classification using the current prediction model. The general paradigm of online learning is as follows. A new observation is aquired and classified by the current prediction model. The model is updated when the label of the new observation is revealed and the update condition is satisfied. The model can be updated after the arrival of each data point (1-by-1), or deferred until a group of data points has arrived (mini-batch by mini-batch).

Among the online learning algorithms introduced in the literature, one of the most popular approach is additive in nature in which given a misclassified observation $(\mathbf{x}_i, y_i)$, the classification model is updated by shifting along the direction of $\mathbf{w} + \alpha_i y_i \mathbf{x}_i \rightarrow \mathbf{w}$ where $\mathbf{w}$ is the weight vector, $y_i \in \{-1,1\}$ is the class label of $\mathbf{x}_i$ and $\alpha_i$ is the weight of misclassified observation. Well-known additive models include the Perceptron [Rosenblatt, 1958], Online Gradient Descent (OGD) [Zinkevich, 2003], Passive Aggressive learning (PA) [Crammer *et al.*, 2006], Soft Confident Weighted (SCW) [Wang *et al.*, 2012], and Adaptive Regularization of Weights (AROW) [Crammer *et al.*, 2009; Crammer *et al.*, 2013]. Ensemble methods have also been proposed for online learning. Online Bagging and Online Boosting [Oza and Russell, 2005] are two well-known online ensemble algorithms. Other algorithms such as the Bayesian-based method [Nguyen *et al.*, 2016a] and the Ellipsoid method [Yang *et al.*, 2009] have also been introduced recently.

Existing approaches have their shortcomings. First, the number of model updates is usually high. For example, in Online Bagging, base classifiers are always updated after the arrival of each new data point. Some algorithms only support 1-by-1 learning but not mini-batch learning. Finally, approaches like the Bayesian-based method [Nguyen *et al.*, 2016a] estimate the distribution of each class and have problems dealing with very high dimensional datasets. Therefore, an algorithm that supports both 1-by-1 and mini-batch learning, only perfoms a small number of updates, and works well for high dimensional datasets, is highly desirable.

In this paper, we propose a novel ensemble-based online learning algorithm for very high dimensional data. To deal with the high dimensional data, our algorithm uses the theory of random projections [Johnson and Lindenstrauss, 1984] to project new observations to low dimension subspaces, thereby obtaining different data schemes for the ensemble of homogenuous base classifiers. Our base classifiers are generated by Naïve Bayes learning, and the final class prediction is obtained by a fix combining rule. In the training process, our algorithm only performs updates when the arrived observations are misclassified, and the update is done in 1-by-1 or mini-batch mode.

The combination between random projections and Naïve Bayes classifier for online learning proposes a novel homogeneous ensemble method to solve the online learning problem. First, when the dimension of data is high, Naïve Bayes classifiers take a long time to train since the likelihood distribution is computed for each feature. By using random projection we first project the input data to low dimensional space and then learned the Naïve Bayes classifiers on the projected data, resulting in the fast learning of Naïve Bayes classifiers. In an online setting we are considering, the observations come one at a time. So at instance $t$, we only have one observation for our ensemble of classifiers. Random projection provides a principled way for us to create a set of 'observations' from one single incoming observation with good diversity for our ensemble of base classifiers. Since random projection is unstable, from one observation, we could create many diverse training data to train the ensemble of homogenous set of classifiers. The ensemble of Naïve Bayes classifiers is expected to obtain better result than a single classifier due to the characteristic of ensemble system [Dietterich, 2000].

The paper is organized as follows. In section 2, we briefly discuss random projection and then develop the ensemble system for online learning based on random projections and Naïve Bayes classifier. Experimental studies are presented in section 3 in which we conduct experiments on thirty two datasets and compare the results of the proposed framework to a number of benchmark algorithms. Our conclusions appears in the last section.

## 2. Proposed method

### 2.1. Random projection

In 1984, Johnson and Lindenstrauss (JL) published a paper about extending Lipschitz continuous maps from metric spaces to Euclidean spaces and introduced the JL Lemma [Johnson and Lindenstrauss, 1984]. The lemma specifies a linear transformation from a $p$-dimensional space $\mathbb{R}^p$ (called up-space) to a $q$-dimensional space $\mathbb{R}^q$ (called down-space). Specifically, given a finite set of $p$-dimensional data vector $\mathcal{D} = \{\mathbf{x}_1, \mathbf{x}_2, \ldots, \mathbf{x}_n\} \subset \mathbb{R}^p$, there exists a linear transformation $\mathrm{T}: \mathbb{R}^p \to \mathbb{R}^q: \mathbf{Z} = \mathrm{T}[\mathcal{D}] = \{\mathbf{z}_1, \mathbf{z}_2, \ldots, \mathbf{z}_n\} \subset \mathbb{R}^q$, $\mathbf{z}_i = \mathrm{T}(\mathbf{x}_i)$ that in probability preserves distance between observations under certain conditions. The linear transformation T can be represented by a matrix $\mathbf{R}$ so that $\mathbf{z}_i = \mathrm{T}(\mathbf{x}_i) = \mathbf{R}\mathbf{x}_i$. When each element of the matrix is generated according to a specified random distribution, T is known as random projection.

Random projection has two desirable properties:

- Random projections are useful in dimensionality reduction since the dimension of down-space is usually much lower than that of up-space i.e. $q < p$. In fact, in some situations, random projection is preferred to Principle Component Analysis (PCA). First, the directions of random projection are independent of the data while those of PCA are data-dependent. This is useful in situations where data cannot be accessed all at once, such as in data streaming. Moreover, generating the principle components is computationally expensive compare to generating the random matrix in random projection [Bingham and Mannila, 2001].

- Fern and Brodley [2003] indicated that random projections are unstable in the sense that the datasets generated from an original data source based on random matrices can be quite different. This property is significant since other sampling methods like bootstrapping only generate slightly different dataset schemes. An ensemble system constructed based on a set of random projections can therefore have a lot of diversity.

In this paper, $K$ random matrices $\mathcal{R} = \{\mathbf{R}^{(k)}\}\ k = 1, \ldots, K$ are generated to construct the ensemble system. We follow the construction of random matrix in [Avogadri and Valentini, 2009] in which the projections are simply obtained by using a $(p \times q)$ random matrix $\mathbf{R}^{(k)} = 1/\sqrt{q} \{r_{ij}^{(k)}\}$ where $r_{ij}$ are random variables such that $\mathrm{E}\left(r_{ij}^{(k)}\right) = 0$ and $\mathrm{var}\left(r_{ij}^{(k)}\right) = 1$. Several forms of $\mathbf{R}^{(k)}$ are summarized as:

- ***Plus-minus-one or Bernoulli random projection:*** $r_{ij}^{(k)}$ is randomly chosen in $\{-1, 1\}$ such that $\mathrm{P}\left(r_{ij}^{(k)} = 1\right) = \mathrm{P}\left(r_{ij}^{(k)} = -1\right) = 1/2$

- ***Achlioptas random projection:*** $r_{ij}^{(k)}$ is randomly chosen in $\{-\sqrt{3}, 0, \sqrt{3}\}$ such that $\mathrm{P}\left(r_{ij}^{(k)} = \sqrt{3}\right) = \mathrm{P}\left(r_{ij}^{(k)} = -\sqrt{3}\right) = 1/6$ and $\mathrm{P}\left(r_{ij}^{(k)} = 0\right) = 2/3$

- **Gaussian random projection:** $r_{ij}^{(k)}$ is distributed according to a Gaussian distribution $\mathcal{N}(0,1)$

## 2.2. Class label prediction

Having $\mathcal{R}$ on hand, the class label of each observation is predicted by using an ensemble of Naïve Bayes classifiers. The Naïve Bayes classifier is a well-known learning algorithm based on Bayes theorem having assumptions about conditional independence between features of observation. Despite the oversimplified assumptions, Naive Bayes classifiers are fast and efficient to train, which is important for streaming data. In detail, given a $p$-dimension vector $\mathbf{x} = (x_1, x_2 \ldots, x_p)$, the posterior probability that $\mathbf{x}$ belongs to class label $y_m$ is given by:

$$P(y_m|\mathbf{x}) = P(y_m|x_1, x_2, \ldots, x_p) \sim P(y_m, x_1, x_2, \ldots, x_p) \sim P(y_m) \prod_{i=1}^{p} P(x_i|y_m) \qquad (1)$$

The likelihood $P(x_i|y_m)$ is computed based on the assumption about the distribution of each feature $x_i$ given $y_m$, such as $x_i|y_m \sim \mathcal{N}(\mu_{mi}, \sigma_{mi}^2)$ in which the parameters $\mu_{mi}$ and $\sigma_{mi}^2$ are computed from the training observations.

When applying this to online classification, at the $t^{th}$ step, the observations in mini-batch $\mathcal{D}_t$ will be projected to the down-space by using each random projection in $\mathcal{R}$. We denoted $\mathbf{Z}_t^k$ as the projection of $\mathcal{D}_t$ associated with the $k^{th}$ projection matrix $\mathbf{R}^{(k)}$

$$\mathbf{Z}_t^{(k)} = \frac{1}{\sqrt{q}} \mathcal{D}_t \mathbf{R}^{(k)} \quad k = 1, \ldots, K \qquad (2)$$

**Assumption 1 (Naïve Bayes):** The features in $\mathbf{z}_t^{(k)}$ are assumed to satisfy the conditional independence assumption.

It has been shown that some violation of the independence assumption of the attributes does not matter [Domingos, Pazzani, 1996]

For each observation $\mathbf{x}_t$ in $\mathcal{D}_t$, denoted its projected vector associated with $\mathbf{R}^{(k)}$ as $\mathbf{z}_t^{(k)} = \{z_{tj}^k\}\ j = 1, \ldots, q$. Based on Assumption 1, the posterior probability that $\mathbf{x}_t$ belong to class label $y_m$ is given by:

$$P(y_m|\mathbf{x}_t) = P\left(y_m|\mathbf{z}_t^{(k)}\right) \sim P(y_m) \prod_{j=1}^{q} P(z_{tj}^k|y_m) \qquad (3)$$

In this paper, we assume that the distribution of likelihood is Gaussian i.e. $z_{tj}^k|y_m \sim \mathcal{N}\left(\mu_{mj}^{(k)}, \sigma_{mj}^{2\,(k)}\right)$. Since the likelihood distribution of each projected attribute is unknown, we used the Gaussian distribution to approximate it. Based on the Central Limit Theorem, Gaussian can be used to approximate a wide range of other distributions such as Poisson, Binominal, and Gamma when we have large enough data [Balakrishnan, Nevzorov, 2003]. Therefore, we have:

$$P(z_{tj}^k|y_m) = \frac{1}{\sqrt{2\pi}\sigma_{mj}^{(k)}} \exp\left\{-\frac{1}{2}\left(\frac{z_{tj}^k - \mu_{mj}^{(k)}}{\sigma_{mj}^{(k)}}\right)^2\right\} \qquad (4)$$

Taking logs, we obtain:

$$\log\left\{P\left(y_m|\mathbf{z}_t^{(k)}\right)\right\} \sim \log\{P(y_m)\} - \sum_{j=1}^{q}\left\{\log\left\{\sqrt{2\pi}\sigma_{mj}^{(k)}\right\} + \frac{1}{2}\left(\frac{z_{tj}^k - \mu_{mj}^{(k)}}{\sigma_{mj}^{(k)}}\right)^2\right\} \qquad (5)$$

The hypotheses from $K$ base classifiers are combined to obtain the final hypothesis. Several popular fixed combining methods, namely Sum, Product, Majority Vote, Max, Min, and Median can be used as the combiner [Nguyen *et al.*, 2014; Nguyen *et al.*, 2016b; Kittler *et al.*, 1998]. Vote and Sum are the most popular rules and have been successfully applied to many combining classifier situations. In this work, we use the Sum rule to combine the outputs of $K$ classifiers:

$$\mathbf{x}_t \in y_s \text{ if } s = \arg\max_{m=1,\ldots,M} \sum_{k=1}^{K} \log\{P(y_m|\mathbf{z}^k)\} \qquad (6)$$

### 2.3. Parameters update

After obtaining the predicted class labels of observations in $\mathcal{D}_t$, we update the parameters of Naïve Bayes classifiers. Theorem 1 provides the equations to update the mean and variance of the likelihood distribution function.

**Theorem 1**: Assume that at the $(t-1)^{th}$ step we have received $(t-1)$ mini-batches $\{\mathcal{D}_i | i = 1, \ldots, t-1\}$ in which mini-batch $\mathcal{D}_i$ has $n_i$ 1-dimensional observations i.e. $\mathcal{D}_i = \{z_{i,1}, z_{i,2}, \ldots, z_{i,n_i}\}$. Denote $\mu(t)$ and $\sigma^2(t)$ as the mean and variance of model at the $t^{th}$ step, the update equations are given by:

$$\mu(t) = \frac{1}{\sum_{i=1}^{t} n_i}\left\{\left(\sum_{i=1}^{t-1} n_i\right)\mu(t-1) + \sum_{i=1}^{n_t} z_{t,i}\right\}$$

$$\sigma^2(t) = \frac{1}{\sum_{i=1}^{t} n_i} \left\{ \left(\sum_{i=1}^{t-1} n_i\right) \left(\sigma^2(t-1) + \left(\mu(t-1) - \mu(t)\right)^2\right) + \sum_{i=1}^{n_t} \left(z_{t,i} - \mu(t)\right)^2 \right\}$$

**Corollary 1**: When a single observation $z_t$ arrives in the sequence, the update equations for $\mu(t)$ and $\sigma^2(t)$ at $t^{th}$ step are given by:

$$\mu(t) = \frac{1}{t}\{(t-1)\mu(t-1) + z_t\}$$

$$\sigma^2(t) = \frac{1}{t}\left\{(t-1)\left(\sigma^2(t-1) + \left(\mu(t-1) - \mu(t)\right)^2\right) + \left(z_t - \mu(t)\right)^2\right\}$$

**Corollary 2**: If the model is updated with every arrived observation, when $t \to N$, $\mu(t) \to \bar{\mu}$ and $\sigma^2(t) \to \sigma^2$ in which $\bar{\mu}$ and $\sigma^2$ are sample mean and variance computed on the whole dataset.

The proof of Theorem1 is given in the Appendix while the proofs of Corollary 1 and 2 are straightforward.

In this study, we use the misclassified observations in $\mathcal{D}_t$ to update the classification model. They are first segmented into $M$ mini-batches $\mathcal{D}_{tm}$ $m = 1, \ldots, M$ in which $\mathcal{D}_{tm}$ contains all the misclassified observations that belong to class label $y_m$ i.e. $\mathbf{x} \in \mathcal{D}_{tm}$ if $y(\mathbf{x}) = y_m$. The mini-batch $\mathcal{D}_{tm}$ will be used to update the mean and variance of the likelihood distribution associated with the $j^{th}$ feature in the down-spaces among all $K$ projections.

$$\mu_{mj}^{(k)}(t) = \frac{1}{\sum_{i=1}^{t} n_{im}} \left\{ \left(\sum_{i=1}^{t-1} n_{im}\right) \mu_{mj}^{(k)}(t-1) + \sum_{i=1}^{n_{tm}} z_{t,ij}^{(k)} \right\} \tag{7}$$

$$\sigma_{mj}^{2\ (k)}(t) = \frac{1}{\sum_{i=1}^{t} n_{im}} \left\{ \left(\sum_{i=1}^{t-1} n_{im}\right) \left(\sigma_{mj}^{2\ (k)}(t-1) + \left(\mu_{mj}^{(k)}(t-1) - \mu_{mj}^{(k)}(t)\right)^2\right) + \sum_{i=1}^{n_{tm}} \left(z_{t,ij}^{(k)} - \mu_{mj}^{(k)}(t)\right)^2 \right\} \tag{8}$$

where $n_{tm} = |\mathcal{D}_{tm}|$, $m = 1, \ldots, M$, $k = 1, \ldots, K$

**Remark**: The online update equations (7) and (8) need only keep track of the cumulative item count in $\mathcal{D}_{tm}$ up to $t$-1, but not the actual item values. So a mini-batch can be discarded once it is used for update.

In case of 1-by-1 training, the model is updated when the predicted label is different from the ground truth class label as follows:

$$\mu_{mj}^{(k)}(t) = \frac{1}{t}\left\{(t-1)\mu_{mj}^{(k)}(t-1) + z_{t,j}^{(k)}\right\} \tag{9}$$

$$\sigma^{2}_{mj}{}^{(k)}(t) = \frac{1}{t}\left\{(t-1)\left(\sigma^{2}_{mj}{}^{(k)}(t-1) + \left(\mu^{(k)}_{mj}(t-1) - \mu^{(k)}_{mj}(t)\right)^{2}\right) + \left(z^{(k)}_{t,j} - \mu^{(k)}_{mj}(t)\right)^{2}\right\} \qquad (10)$$

We have the following algorithm for mini-batch online training in general.

**Algorithm: Ensemble online training based on random projection and Naïve Bayes**

**1. Parameter initialization**

Input: $\mu^{(k)}_{mj}(0)$ and $\sigma^{2}_{mj}{}^{(k)}(0)$, ensemble size $K$,

    down-space dimension $q$

For $k = 1 \ldots K$

    For $m = 1 \ldots M$

        For $j = 1 \ldots q$

            Set $\mu^{(k)}_{mj} = \mu^{(k)}_{mj}(0), \sigma^{2}_{mj}{}^{(k)} = \sigma^{2}_{mj}{}^{(k)}(0)$

        End

    End

End

**2. Random matrix generation**

For $k = 1 \ldots K$

    Generate $\mathbf{R}^{k} = \{r^{k}_{ij}\}, \; r^{k}_{ij} \sim \mathcal{N}(0,1)$

End

**3. Class label prediction**

Input: $\mathcal{D}_t = \{(\mathbf{x}_t, y(\mathbf{x}_t))\}$

For $k = 1, \ldots, K$

    $\mathbf{Z}^{(k)}_t = \frac{1}{\sqrt{q}} \mathcal{D}_t \mathbf{R}^{(k)}$

    Compute $\log\{P(y_m | \mathbf{z}^{(k)}_t)\}$ using (5)

```
End
Predict label ŷ(x_t) using Sum rule(6)
```
**4. Parameter update**

```
Partition misclassified observations from 𝒟_t into 𝒟_tm such that x ∈ 𝒟_tm if
y(x) = y_m
For  k = 1 … K
   For  m = 1 … M
      For  j = 1 … q
         Update μ_{mj}^{(k)}(t) using (7)
         Update σ²_{mj}^{(k)}(t) using (8)
      End
   End
End
```

## 3. Empirical Studies

### 3.1. Setup

To evaluate the performance of the proposed method, we carry out experiments on thirty two UCI labeled datasets [UCI] and one labeled synthetic dataset named GM. The GM dataset consists of 1000 observations generated from a Gaussian Mixture with 3 components in equal proportions. The means of the components are $\{1/2, \ldots, 1/2\}_{1000}$, $\{0, \ldots, 0\}_{1000}$, and $\{-1/2, \ldots, -1/2\}_{1000}$ respectively while the corresponding standard deviations are $\text{diag}\{1, \ldots, 1\}_{1000}$, $\text{diag}\{2, \ldots, 2\}_{1000}$, and $\text{diag}\{3, \ldots, 3\}_{1000}$ The information about the datasets is shown in Table 1.

We perform extensive comparative studies with a number of state-of-the-art algorithms as benchmarks: PA [Crammer *et al.*, 2006], SCW [Wang *et al.*, 2012], OGD [Zinkevich, 2003], AROW [Crammer *et al.*, 2009; Crammer *et al.*, 2013] (we use the implementation in LIBOL library [Hoi *et al.*, 2014] for these algorithms, default value for parameters are used if available), and Online Bagging [Oza and Russell, 2005] (we use the implementation in MOA library [Bifet *et al.*, 2010]). AROW, OGD, and SCW are algorithms published in top machine learning venues like NIPS and ICML. Online Bagging is a high performance ensemble online learning method. For the proposed method, Gaussian random projection is used to generate the random matrix. The number of learners in Online Bagging and *K* in the proposed method are set to 200 as in [Nguyen *et al.*, 2016b], and the dimension of all down-spaces is set to $q = 2\log_2 p$. The parameters for Naïve Bayes classifiers are simply initialized as $\mu_{mj}^{(k)}(0) = 0$ and $\sigma_{mj}^{2\ (k)}(0) = 1$ for $k = 1 \ldots K, m = 1 \ldots M$.

In this study, the proposed method uses 1-by-1 learning, denoted by RPNB(1b1), since all benchmark algorithms use 1-by-1 learning. The proposed method is compared to the benchmark algorithms with respect to the error rate and F1 score (which is the harmonic mean of Precision and Recall) [Sokolova and Lapalme, 2009]. We draw 10 random permutations from each data to obtain the sequences of arriving data, run the test, and compute the average of the 10 classification error rates, F1 Scores, and number of updates. Here we followed the performance measurements from LIBOL library [Hoi et al., 2014] where the authors used criteria such as mistake rate (classification error rate), and the number of updates (to measure the model stability) to evaluate the performance. In this paper, we conducted Wilcoxon signed rank test [Demsar, 2006] (level of significance is 0.05) to compare a pairs of algorithms, i.e. a benchmark algorithm and the proposed algorithm. Here we tested the specific null hypothesis that "two methods perform equally". Based on the value of statistic in Wilcoxon procedure, we could obtain the P-Value of the test. The performance scores of two methods are treated as significantly different if the P-Value of the test is smaller than a given significant level α. When the test indicates that the performance of two algorithms is different, we then use the classification error rate to decide which algorithm wins on a particular dataset and count the number of wins and losses on the set of datasets.

**3.2. Results and Discussions**

The mean and variance of error rates and F1 Score of the benchmark algorithms and the proposed method are shown in Tables 2 and 3. The statistical test results in Figure 1 show that RPNB(1b1) significantly outperformed all benchmark algorithms with respect to error rate . Comparing to AROW, we rejected 27 null hypotheses, in which RPNB(1b1) is better than AROW on 24 datasets and worse on 3. Comparing with Online Bagging, we rejected 27 null hypothesis, in which RPNB(1b1) is better on 22 datasets and worse on 5. RPNB(1b1) is also significantly better than PA (32 wins, 0 loss), OGD (32 wins and 1 loss), and SCW (25 wins and 3 losses). For F1 score, the statistical test results in Figure 2 show that RPNB(1b1) significantly outperform all benchmark algorithms.

| File name | # of features | # of observations | # of classes |
|---|---|---|---|
| Breast Cancer | 9 | 683 | 2 |
| Breast Tissue | 9 | 106 | 6 |
| Chess-krvk | 6 | 28056 | 18 |
| Conn Bench Vowel | 10 | 528 | 11 |
| Contraceptive | 9 | 1473 | 3 |
| Ecoli | 7 | 336 | 8 |
| GM | 1000 | 1000 | 3 |
| Hayes Roth | 4 | 160 | 3 |
| Ionosphere | 34 | 351 | 2 |
| Iris | 4 | 150 | 3 |
| Isolet | 617 | 7797 | 26 |
| Led7digit | 7 | 500 | 10 |
| Letter | 16 | 20000 | 26 |
| Madelon | 500 | 2000 | 2 |
| Marketing | 13 | 6876 | 9 |
| Monk-2 | 6 | 432 | 2 |
| Multiple Features | 649 | 2000 | 10 |
| Musk1 | 166 | 476 | 2 |
| Musk2 | 166 | 6598 | 2 |
| Nursery | 8 | 12960 | 5 |
| Optdigits | 64 | 5620 | 10 |
| Optical | 64 | 3823 | 10 |
| Penbased | 16 | 10992 | 10 |
| Satimage | 36 | 6435 | 6 |
| Skin_NonSkin | 3 | 245057 | 2 |
| Soybean | 35 | 307 | 19 |
| Tae | 5 | 151 | 3 |
| Tic_Tac_Toe | 9 | 958 | 2 |
| Twonorm | 20 | 7400 | 2 |
| Vertebral | 6 | 310 | 3 |
| Waveform_w_Noise | 40 | 5000 | 3 |
| Waveform_wo_Noise | 21 | 5000 | 3 |
| Zoo | 16 | 101 | 7 |

TABLE.1. INFORMATION OF DATASETS IN EVALUATION

Figure 3 shows the number of times the model is updated on two datasets. Clearly, the proposed method requires significantly less number of updates than the benchmark algorithms. For example, on GM datasets, the average number of updates of RPNB(1b1) is 29.6, nearly 9 times less than that of OGD algorithm (247.4) which has the second smallest number of updates among all benchmark algorithms.

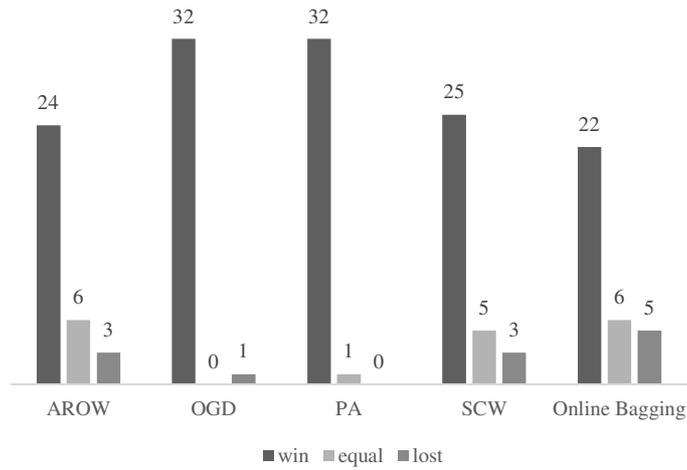

Figure 1: Statistical test result comparing RPNB(1b1) to the benchmark algorithms with respect to error rate

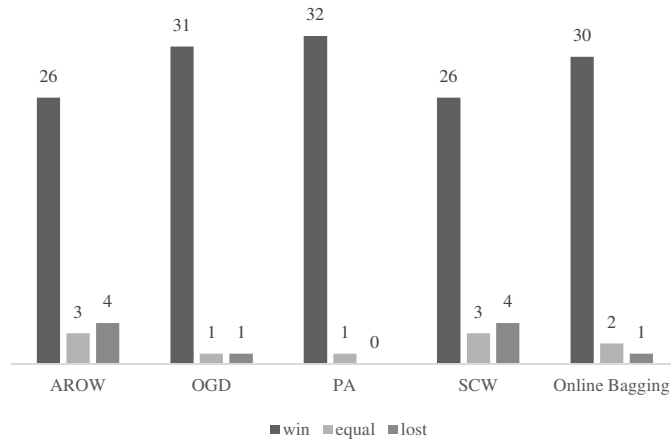

Figure 2: Statistical test result comparing RPNB(1b1) to the benchmark algorithms with respect to F1 score

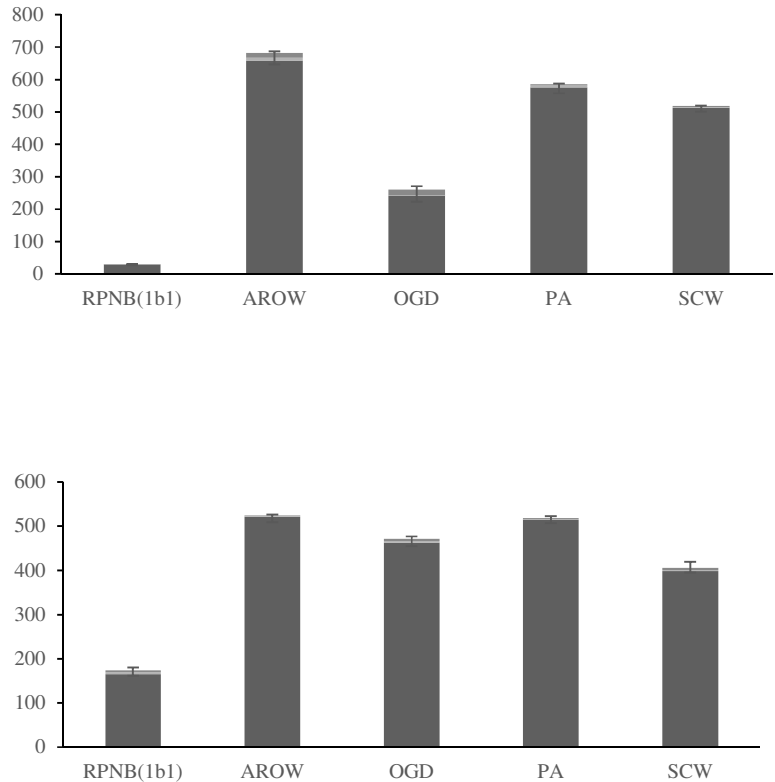

Figure 3: Number of updates on GM (top figure) and Conn Bench Vowel (bottom figure) dataset.

The comparative study has shown that our algorithm comfortably outperforms the state-of-the-art benchmark algorithms on the datasets used here. We believe that the success of RPNB(1b1) is partially due to the diverse data schemes generated by random projection used in our ensemble. In addition, we used Naïve Bayes, a simple but efficacious learning algorithm [Webb *et al.*, 2005] to generate the base classifiers where the parameter updates are simple and fast to compute. For fast data streams, our algorithm can be run in minibatch mode which would further reduce the number of parameter updates.

Our use of random projection in this work serves two purposes: dimensionality reduction if the feature dimension is high, and more importantly, generation of diverse data schemes for the ensemble. A significant departure from the original JL Lemma is that the down-space dimension $q$ is a function of number of observations in JL, whereas in this work the down-space dimension is computed as a function of feature dimension $p$. In other words, our dimensionality reduction step is inspired by the JL lemma, but there is no probabilistic guarantee about distance preservation in our approach. Our down-space dimension is the same as $p$ when $p < 5$. In this case, it is only the diverse data schemes we are interested in when applying random projection.

## 4. Conclusions

In this paper, we have introduced an ensemble-based online learning algorithm using random projection and Naïve Bayes classifiers. In our approach, the parameters of the Naïve Bayes classifiers are simply initialized at the beginning and then updated if arrived observations are misclassified. We proposed the update equations for Naïve Bayes classifiers' parameters via mini-batch by mini-batch learning and 1-by-1 learning. Extensive experimental results for the 1 by 1 case demonstrated the benefit of our approach compared with several well-known benchmark algorithms with respect to classification error rate, F1 score, and the number of updates.

## A. Appendix: Proof of Theorem 1

$$\mu(t) = \frac{1}{\sum_{i=1}^{t} n_i} \left\{ \sum_{j=1}^{t} \sum_{i=1}^{n_i} z_{j,i} \right\} = \frac{1}{\sum_{i=1}^{t} n_i} \left\{ \sum_{j=1}^{t-1} \sum_{i=1}^{n_i} z_{j,i} + \sum_{i=1}^{n_t} z_{t,i} \right\} = \frac{1}{\sum_{i=1}^{t} n_i} \left\{ \left( \sum_{i=1}^{t-1} n_i \right) \mu(t-1) + \sum_{i=1}^{n_t} z_{t,i} \right\}$$

$$\sigma^2(t) = \frac{1}{\sum_{i=1}^{t} n_i} \left\{ \sum_{j=1}^{t} \sum_{i=1}^{n_i} \left( z_{j,i} - \mu(t) \right)^2 \right\} = \frac{1}{\sum_{i=1}^{t} n_i} \left\{ \sum_{j=1}^{t-1} \sum_{i=1}^{n_i} \left( z_{j,i} - \mu(t) \right)^2 + \sum_{i=1}^{n_t} \left( z_{t,i} - \mu(t) \right)^2 \right\}$$

$$= \frac{1}{\sum_{i=1}^{t} n_i} \left\{ \sum_{j=1}^{t-1} \sum_{i=1}^{n_i} \left( z_{j,i} - \mu(t-1) + \mu(t-1) - \mu(t) \right)^2 + \sum_{i=1}^{n_t} \left( z_{t,i} - \mu(t) \right)^2 \right\}$$

$$= \frac{1}{\sum_{i=1}^{t} n_i} \left\{ \sum_{j=1}^{t-1} \sum_{i=1}^{n_i} \left( z_{j,i} - \mu(t-1) \right)^2 + \sum_{j=1}^{t-1} \sum_{i=1}^{n_i} \left( \mu(t-1) - \mu(t) \right)^2 \right.$$

$$\left. + \sum_{j=1}^{t-1} \sum_{i=1}^{n_i} 2 \left( z_{j,i} - \mu(t-1) \right) \left( \mu(t-1) - \mu(t) \right) + \sum_{i=1}^{n_t} \left( z_{t,i} - \mu(t) \right)^2 \right\}$$

$$= \frac{1}{\sum_{i=1}^{t} n_i} \left\{ \left( \sum_{i=1}^{t-1} n_i \right) \left( \sigma_{t-1}^2 + (\mu(t-1) - \mu(t))^2 \right) + 2(\mu_{t-1} - \mu_t) \sum_{j=1}^{t-1} \sum_{i=1}^{n_i} (z_{j,i} - \mu(t-1)) \right.$$
$$\left. + \sum_{i=1}^{n_t} (z_{t,i} - \mu(t))^2 \right\}$$

Since

$$2(\mu(t-1) - \mu(t)) \sum_{j=1}^{t-1} \sum_{i=1}^{n_i} (z_{j,i} - \mu(t-1)) = 2(\mu(t-1) - \mu(t)) \left\{ \sum_{j=1}^{t-1} \sum_{i=1}^{n_i} z_{j,i} - \left( \sum_{i=1}^{t-1} n_i \right) \mu(t-1) \right\}$$

$$= 2(\mu(t-1) - \mu(t)) \left\{ \left( \sum_{i=1}^{t-1} n_i \right) \mu(t-1) - \left( \sum_{i=1}^{t-1} n_i \right) \mu(t-1) \right\} = 0$$

We have

$$\sigma^2(t) = \frac{1}{\sum_{i=1}^{t} n_i} \left\{ \left( \sum_{i=1}^{t-1} n_i \right) \left( \sigma^2(t-1) + (\mu(t-1) - \mu(t))^2 \right) + \sum_{i=1}^{n_t} (z_{t,i} - \mu(t))^2 \right\}$$

|  | RPNB(1b1) |  | AROW |  | OGD |  | PA |  | SCW |  | Online Bagging |  |
|---|---|---|---|---|---|---|---|---|---|---|---|---|
|  | Mean | Var | Mean | Var | Mean | Var | Mean | Var | Mean | Var | Mean | Var |
| Breast Cancer | 0.0490 | 5.25E-06 | 0.1476 ↑ | 4.02E-05 | 0.2004 ↑ | 1.64E-04 | 0.2045 ↑ | 5.83E-05 | 0.1690 ↑ | 8.33E-05 | 0.0479 | 1.95E-05 |
| Breast Tissue | 0.5123 | 4.28E-04 | 0.5745 ↑ | 3.00E-03 | 0.7915 ↑ | 1.25E-03 | 0.7811 ↑ | 1.22E-03 | 0.6321 ↑ | 6.73E-03 | 0.4513 ↓ | 8.56E-04 |
| Chess-krvk | 0.7038 | 1.16E-05 | 0.8110 ↑ | 8.16E-05 | 0.8298 ↑ | 1.03E-05 | 0.8524 ↑ | 3.43E-06 | 0.8271 ↑ | 1.87E-04 | 0.6974 ↓ | 4.03E-06 |
| Conn Bench Vowel | 0.3216 | 9.81E-05 | 0.6203 ↑ | 3.73E-04 | 0.7187 ↑ | 3.17E-04 | 0.7648 ↑ | 2.03E-04 | 0.5983 ↑ | 3.10E-04 | 0.4065 ↑ | 2.60E-04 |
| Contraceptive | 0.5133 | 1.28E-04 | 0.5119 | 5.02E-05 | 0.6000 ↑ | 1.42E-04 | 0.6356 ↑ | 1.40E-04 | 0.5136 | 3.38E-05 | 0.5315 ↑ | 6.83E-05 |
| Ecoli | 0.2774 | 1.25E-04 | 0.3449 ↑ | 2.52E-03 | 0.4551 ↑ | 2.74E-04 | 0.5018 ↑ | 2.55E-04 | 0.3655 ↑ | 9.72E-04 | 0.2755 | 2.20E-04 |
| GM | 0.0296 | 1.44E-06 | 0.3379 ↑ | 1.78E-04 | 0.2460 ↑ | 2.35E-04 | 0.2316 ↑ | 1.38E-05 | 0.2290 ↑ | 2.60E-05 | 0.6700 ↑ | 2.34E-05 |
| Hayes Roth | 0.3675 | 5.22E-04 | 0.6750 ↑ | 1.45E-03 | 0.6425 ↑ | 6.00E-04 | 0.6431 ↑ | 1.34E-03 | 0.6538 ↑ | 1.13E-03 | 0.3900 | 8.83E-04 |
| Ionosphere | 0.1134 | 5.97E-05 | 0.1735 ↑ | 5.75E-05 | 0.1972 ↑ | 1.54E-04 | 0.2254 ↑ | 9.98E-05 | 0.1889 ↑ | 1.20E-04 | 0.1810 ↑ | 4.44E-04 |
| Iris | 0.0727 | 6.62E-05 | 0.1240 ↑ | 8.11E-04 | 0.3493 ↑ | 5.80E-04 | 0.3953 ↑ | 1.30E-03 | 0.1233 ↑ | 3.22E-04 | 0.1037 ↑ | 1.87E-04 |
| Isolet | 0.1129 | 6.05E-06 | 0.1446 ↑ | 1.73E-05 | 0.1706 ↑ | 1.52E-05 | 0.1755 ↑ | 6.05E-06 | 0.0785 ↓ | 2.11E-06 | 0.1989 ↑ | 2.12E-05 |
| Led7digit | 0.3188 | 4.74E-05 | 0.3278 | 1.45E-04 | 0.4128 ↑ | 1.63E-04 | 0.5302 ↑ | 4.64E-04 | 0.3356 ↑ | 6.22E-05 | 0.3418 ↑ | 1.23E-04 |
| Letter | 0.2830 | 7.19E-06 | 0.4687 ↑ | 4.91E-04 | 0.4372 ↑ | 9.97E-06 | 0.5309 ↑ | 1.64E-06 | 0.4859 ↑ | 7.97E-04 | 0.3652 ↑ | 9.34E-06 |
| Madelon | 0.4334 | 1.44E-05 | 0.4795 ↑ | 8.15E-05 | 0.4778 ↑ | 1.52E-04 | 0.5020 ↑ | 1.53E-04 | 0.4715 ↑ | 1.47E-04 | 0.5007 ↑ | 1.85E-06 |
| Marketing | 0.6962 | 1.11E-05 | 0.7148 ↑ | 1.25E-04 | 0.7566 ↑ | 1.35E-05 | 0.7840 ↑ | 3.08E-05 | 0.7324 ↑ | 1.14E-04 | 0.6961 | 9.37E-06 |
| Monk-2 | 0.0845 | 1.24E-04 | 0.2537 ↑ | 1.10E-04 | 0.3157 ↑ | 5.59E-05 | 0.3718 ↑ | 9.45E-05 | 0.2632 ↑ | 1.07E-04 | 0.1086 ↑ | 2.94E-04 |
| Multiple Features | 0.1980 | 6.00E-05 | 0.1481 ↓ | 6.97E-04 | 0.4280 ↑ | 1.06E-04 | 0.5615 ↑ | 1.36E-04 | 0.0919 ↓ | 2.39E-04 | 0.9137 ↑ | 1.85E-05 |
| Musk1 | 0.2193 | 2.09E-04 | 0.3172 ↑ | 2.95E-04 | 0.3105 ↑ | 3.09E-04 | 0.3349 ↑ | 5.70E-04 | 0.2431 ↑ | 1.43E-04 | 0.5198 ↑ | 4.50E-04 |
| Musk2 | 0.0646 | 4.78E-06 | 0.0704 ↑ | 2.90E-06 | 0.1097 ↑ | 5.18E-06 | 0.1196 ↑ | 5.05E-06 | 0.0930 ↑ | 3.14E-06 | 0.1539 ↑ | 1.73E-05 |
| Nursery | 0.1237 | 2.48E-05 | 0.2464 ↑ | 8.88E-06 | 0.2950 ↑ | 3.65E-06 | 0.3829 ↑ | 1.70E-05 | 0.2647 ↑ | 8.37E-05 | 0.0934 ↓ | 2.17E-06 |
| Optdigits | 0.0480 | 2.17E-06 | 0.1248 ↑ | 2.04E-04 | 0.0981 ↑ | 3.57E-06 | 0.0940 ↑ | 5.99E-06 | 0.0632 ↑ | 5.25E-06 | 0.1140 ↑ | 5.58E-06 |
| Optical | 0.0530 | 2.77E-06 | 0.1358 ↑ | 1.68E-04 | 0.1073 ↑ | 6.84E-06 | 0.1044 ↑ | 6.62E-06 | 0.0648 ↑ | 6.93E-06 | 0.1183 ↑ | 5.41E-06 |
| Penbased | 0.0944 | 3.64E-05 | 0.1851 ↑ | 1.28E-03 | 0.1452 ↑ | 8.69E-06 | 0.1773 ↑ | 6.50E-06 | 0.1405 ↑ | 4.17E-04 | 0.1244 ↑ | 1.98E-05 |
| Satimage | 0.1575 | 6.99E-06 | 0.3379 ↑ | 1.76E-04 | 0.3991 ↑ | 6.76E-05 | 0.4652 ↑ | 1.34E-05 | 0.2713 ↑ | 3.86E-04 | 0.2069 ↑ | 2.66E-06 |
| Skin_NonSkin | 0.0029 | 3.32E-07 | 0.0917 ↑ | 2.00E-07 | 0.1636 ↑ | 1.71E-07 | 0.1553 ↑ | 4.52E-07 | 0.0679 ↑ | 3.14E-08 | 0.0152 ↑ | 1.51E-06 |
| Soybean | 0.2596 | 2.77E-05 | 0.2489 | 3.61E-04 | 0.4909 ↑ | 2.14E-04 | 0.5603 ↑ | 4.20E-04 | 0.2873 ↑ | 4.43E-04 | 0.3288 ↑ | 2.17E-04 |
| Tae | 0.6166 | 3.99E-04 | 0.6232 | 7.76E-04 | 0.6517 ↑ | 1.19E-03 | 0.6536 ↑ | 7.55E-04 | 0.6146 | 8.75E-04 | 0.5189 ↓ | 3.66E-04 |

| | | | | | | | | | | | |
|---|---|---|---|---|---|---|---|---|---|---|---|
| Tic_Tac_Toe | 0.3086 | 2.46E-04 | 0.3399 ↑ | 3.99E-05 | 0.4023 ↑ | 2.99E-04 | 0.4410 ↑ | 2.52E-04 | 0.3574 ↑ | 1.33E-04 | 0.3243 | 1.20E-04 |
| Twonorm | 0.0371 | 1.52E-06 | 0.0246 ↓ | 5.70E-07 | 0.0259 ↓ | 7.53E-07 | 0.0365 | 1.03E-06 | 0.0281 ↓ | 6.53E-07 | 0.0253 ↓ | 8.45E-07 |
| Vertebral | 0.2171 | 1.29E-04 | 0.2403 ↑ | 4.79E-04 | 0.3103 ↑ | 3.51E-04 | 0.3632 ↑ | 3.11E-04 | 0.2432 ↑ | 2.36E-04 | 0.2092 | 2.05E-04 |
| Waveform_w_Noise | 0.1647 | 1.26E-05 | 0.1685 | 2.13E-05 | 0.1720 ↑ | 1.09E-05 | 0.2003 ↑ | 1.98E-05 | 0.1688 | 8.94E-06 | 0.2024 ↑ | 7.14E-06 |
| Waveform_wo_Noise | 0.1629 | 1.42E-05 | 0.1601 | 2.86E-05 | 0.1699 ↑ | 3.13E-05 | 0.2061 ↑ | 1.83E-05 | 0.1604 | 1.45E-05 | 0.1924 ↑ | 4.66E-06 |
| Zoo | 0.2020 | 1.61E-04 | 0.1564 ↓ | 9.41E-05 | 0.2861 ↑ | 2.24E-04 | 0.2931 ↑ | 3.96E-04 | 0.1931 | 8.33E-05 | 0.2649 ↑ | 6.67E-04 |

TABLE 2: MEAN AND VARIANCE OF ERROR RATES OF BENCHMARK ALGORITHMS AND PROPOSED METHOD

| | RPNB(1b1) | | AROW | | OGD | | PA | | SCW | | Online Bagging | |
|---|---|---|---|---|---|---|---|---|---|---|---|---|
| | Mean | Var | Mean | Var | Mean | Var | Mean | Var | Mean | Var | Mean | Var |
| Breast Cancer | 0.9460 | 6.42E-06 | 0.8373 ↑ | 5.71E-05 | 0.7903 ↑ | 1.64E-04 | 0.7564 ↑ | 8.15E-05 | 0.8202 ↑ | 8.37E-05 | 0.9482 | 2.15E-05 |
| Breast Tissue | 0.4603 | 4.18E-04 | 0.3753 ↑ | 3.45E-03 | 0.1390 ↑ | 1.00E-03 | 0.2102 ↑ | 1.14E-03 | 0.3564 ↑ | 6.75E-03 | 0.2673 ↑ | 5.26E-04 |
| Chess-krvk | 0.2217 | 1.08E-05 | 0.1427 ↑ | 7.35E-05 | 0.1393 ↑ | 2.00E-05 | 0.1145 ↑ | 4.71E-06 | 0.1050 ↑ | 6.25E-05 | 0.0614 ↑ | 2.09E-06 |
| Conn Bench Vowel | 0.6798 | 9.55E-05 | 0.3701 | 3.47E-04 | 0.2798 ↑ | 3.20E-04 | 0.2340 ↑ | 1.93E-04 | 0.3919 ↑ | 3.86E-04 | 0.2292 ↑ | 2.99E-04 |
| Contraceptive | 0.4786 | 1.01E-04 | 0.4522 ↑ | 7.59E-05 | 0.3792 ↑ | 1.65E-04 | 0.3443 ↑ | 1.68E-04 | 0.4591 ↑ | 3.78E-05 | 0.3801 ↑ | 4.59E-05 |
| Ecoli | 0.4800 | 1.68E-04 | 0.4156 ↑ | 2.14E-03 | 0.2962 ↑ | 1.04E-03 | 0.2471 ↑ | 4.43E-04 | 0.4045 ↑ | 1.34E-03 | 0.2267 ↑ | 1.47E-04 |
| GM | 0.9700 | 1.57E-06 | 0.6409 ↑ | 1.91E-04 | 0.7133 ↑ | 6.25E-04 | 0.7299 ↑ | 3.43E-05 | 0.7330 ↑ | 5.35E-05 | 0.2024 ↑ | 7.21E-04 |
| Hayes Roth | 0.6621 | 5.40E-04 | 0.3114 ↑ | 1.57E-03 | 0.3387 ↑ | 7.81E-04 | 0.3203 ↑ | 1.59E-03 | 0.3248 ↑ | 1.18E-03 | 0.4488 ↑ | 1.03E-03 |
| Ionosphere | 0.8725 | 9.11E-05 | 0.7890 ↑ | 1.19E-04 | 0.7631 ↑ | 2.36E-04 | 0.7423 ↑ | 1.40E-04 | 0.7815 ↑ | 1.55E-04 | 0.8092 ↑ | 4.14E-04 |
| Iris | 0.9274 | 6.66E-05 | 0.8754 ↑ | 8.31E-04 | 0.6513 ↑ | 6.02E-04 | 0.6007 ↑ | 1.43E-03 | 0.8763 ↑ | 3.27E-04 | 0.8679 ↑ | 3.63E-04 |
| Isolet | 0.8870 | 5.93E-06 | 0.8553 ↑ | 1.73E-05 | 0.8293 ↑ | 1.54E-05 | 0.8238 ↑ | 6.15E-06 | 0.9214 ↓ | 2.02E-06 | 0.2043 ↑ | 6.64E-05 |
| Led7digit | 0.6816 | 4.95E-05 | 0.6670 ↑ | 1.61E-04 | 0.5738 ↑ | 2.25E-04 | 0.4482 ↑ | 9.32E-04 | 0.6622 ↑ | 5.94E-05 | 0.3060 ↑ | 3.48E-04 |
| Letter | 0.7124 | 8.77E-06 | 0.5193 ↑ | 5.15E-05 | 0.5614 ↑ | 1.00E-05 | 0.4644 ↑ | 1.94E-06 | 0.5057 ↑ | 8.33E-04 | 0.1358 ↑ | 6.25E-06 |
| Madelon | 0.5666 | 1.44E-05 | 0.5205 ↑ | 8.13E-05 | 0.5222 ↑ | 1.52E-04 | 0.4980 ↑ | 1.53E-04 | 0.5285 ↑ | 1.47E-04 | 0.3357 ↑ | 3.95E-06 |
| Marketing | 0.2454 | 1.32E-05 | 0.2277 ↑ | 6.19E-05 | 0.2011 ↑ | 1.60E-05 | 0.1827 ↑ | 2.65E-05 | 0.2230 ↑ | 5.27E-05 | 0.0883 ↑ | 2.39E-06 |
| Monk-2 | 0.9152 | 1.25E-04 | 0.7461 ↑ | 1.07E-04 | 0.6839 ↑ | 5.63E-05 | 0.6254 ↑ | 1.05E-04 | 0.7366 ↑ | 1.09E-04 | 0.8912 ↑ | 2.96E-04 |
| Multiple Features | 0.8028 | 5.66E-05 | 0.8523 ↓ | 6.86E-04 | 0.5726 ↑ | 1.08E-04 | 0.4362 ↑ | 1.36E-04 | 0.9082 ↓ | 2.35E-04 | 0.0195 ↑ | 7.53E-06 |
| Musk1 | 0.7780 | 2.12E-04 | 0.6817 ↑ | 2.89E-04 | 0.6834 ↑ | 3.16E-04 | 0.6602 ↑ | 6.03E-04 | 0.7537 ↑ | 1.43E-04 | 0.4367 ↑ | 3.01E-03 |
| Musk2 | 0.8831 | 1.21E-05 | 0.8613 ↑ | 8.25E-06 | 0.7891 ↑ | 1.77E-05 | 0.7690 ↑ | 1.61E-05 | 0.8264 ↑ | 1.23E-05 | 0.4802 ↑ | 1.12E-03 |
| Nursery | 0.6747 | 3.07E-04 | 0.4534 ↑ | 3.80E-06 | 0.4276 ↑ | 1.86E-06 | 0.3928 ↑ | 1.25E-05 | 0.4597 ↑ | 1.24E-04 | 0.5354 ↑ | 4.09E-05 |
| Optdigits | 0.9522 | 2.16E-06 | 0.8750 ↑ | 2.07E-04 | 0.9020 ↑ | 3.59E-06 | 0.9058 ↑ | 6.03E-06 | 0.9368 ↑ | 5.27E-06 | 0.6524 ↑ | 8.39E-05 |
| Optical | 0.9473 | 2.81E-06 | 0.8642 ↑ | 1.69E-04 | 0.8929 ↑ | 6.61E-06 | 0.8954 ↑ | 6.45E-06 | 0.9352 ↑ | 6.94E-06 | 0.5982 ↑ | 1.11E-04 |
| Penbased | 0.9010 | 4.62E-05 | 0.8136 ↑ | 1.29E-03 | 0.8541 ↑ | 8.57E-06 | 0.8209 ↑ | 6.68E-06 | 0.8583 ↑ | 4.47E-04 | 0.5595 ↑ | 9.61E-05 |
| Satimage | 0.8150 | 8.38E-06 | 0.6052 ↑ | 2.66E-04 | 0.5773 ↑ | 6.57E-05 | 0.5113 ↑ | 1.52E-05 | 0.6891 ↑ | 4.49E-04 | 0.5152 ↑ | 8.02E-06 |
| Skin_NonSkin | 0.9956 | 7.84E-07 | 0.8678 ↑ | 5.70E-07 | 0.7621 ↑ | 3.39E-07 | 0.7506 ↑ | 1.56E-06 | 0.9057 ↑ | 5.41E-08 | 0.9773 ↑ | 3.40E-06 |
| Soybean | 0.7080 | 1.38E-04 | 0.7137 | 5.40E-04 | 0.3993 ↑ | 5.84E-04 | 0.3330 ↑ | 8.77E-04 | 0.6629 ↑ | 4.73E-04 | 0.1454 ↑ | 4.56E-04 |
| Tae | 0.3748 | 3.64E-04 | 0.3744 | 8.87E-04 | 0.3485 ↑ | 1.21E-03 | 0.3463 ↑ | 7.76E-04 | 0.3845 | 9.59E-04 | 0.3855 | 4.69E-04 |
| Tic_Tac_Toe | 0.6758 | 2.36E-04 | 0.5428 ↑ | 2.08E-04 | 0.5306 ↑ | 4.17E-04 | 0.5148 ↑ | 3.03E-04 | 0.5115 ↑ | 5.65E-04 | 0.4950 ↑ | 1.28E-03 |
| Twonorm | 0.9629 | 1.52E-06 | 0.9754 ↓ | 5.70E-07 | 0.9741 ↓ | 7.52E-07 | 0.9635 | 1.03E-06 | 0.9719 ↓ | 6.53E-07 | 0.9747 ↓ | 8.45E-07 |
| Vertebral | 0.7352 | 2.48E-04 | 0.6958 ↑ | 4.77E-04 | 0.6118 ↑ | 6.37E-04 | 0.5640 ↑ | 4.47E-04 | 0.6936 ↑ | 3.61E-04 | 0.6808 ↑ | 5.03E-04 |
| Waveform_w_Noise | 0.8323 | 1.42E-05 | 0.8314 | 2.13E-05 | 0.8281 ↑ | 1.07E-05 | 0.7999 ↑ | 1.97E-05 | 0.8312 | 8.97E-06 | 0.7154 ↑ | 1.29E-05 |
| Waveform_wo_Noise | 0.8327 | 1.77E-05 | 0.8394 ↓ | 2.88E-05 | 0.8299 | 3.14E-05 | 0.7938 ↑ | 1.83E-05 | 0.8393 ↓ | 1.49E-05 | 0.7175 ↑ | 1.02E-05 |
| Zoo | 0.6485 | 6.93E-04 | 0.7117 ↓ | 1.04E-03 | 0.5031 ↑ | 1.56E-03 | 0.4969 ↑ | 1.02E-03 | 0.6559 | 3.86E-04 | 0.2710 ↑ | 7.13E-04 |

TABLE 3: MEAN AND VARIANCE OF F1 SCORES OF BENCHMARK ALGORITHMS AND PROPOSED METHOD

↓: *The benchmark algorithm is better than RPNB(1b1)*, ↑: *The benchmark algorithm is worse than RPNB(1b1)*

# References

[Nguyen *et al.*, 2016a] T.T.T. Nguyen, T.T. Nguyen, X.C. Pham, A.W.-C. Liew, Y. Hu, T. Liang, C.-T. Li. A novel online Bayes classifier. *in Proceedings of the International Conference on Digital Image Computing: Techniques and Applications (DICTA)*, Gold Coast, Australia, 2016.

[Rosenblatt, 1958] F. Rosenblatt. The perceptron: A probabilistic model for information storage and organization in the brain. *Psychological Review*, 65(6): 386–408, 1958.

[Zinkevich, 2003] M. Zinkevich. Online convex programming and generalized infinitesimal gradient ascent. *in Proceedings of the International Conference on Machine Learning (ICML)*, pages 928–936, 2003.

[Crammer *et al.*, 2006] K. Crammer, O. Dekel, J. Keshet, S. Shalev-Shwartz, Y. Singer. Online passive aggressive algorithms. *Journal of Machine Learning Research*, 7: 551–585, 2006.

[Wang et al., 2012] J. Wang, P. Zhao, S.C.H Hoi. Exact soft confidence-weighted learning. *in Proceedings of the 29th International Conference on Machine Learning (ICML)*, Edinburgh, Scotland, UK, 2012.

[Crammer *et al.*, 2009] K. Crammer, A. Kulesza, M. Dredze. Adaptive regularization of weight vectors. *in Proceedings of the 22th Advances in Neural Information Processing Systems (NIPS)*, pages 414–422, 2009.

[Crammer *et al.*, 2013] K. Crammer, A. Kulesza, M. Dredze. Adaptive regularization of weight vectors. *Machine Learning*, 91(2): 155–187, 2013.

[Oza and Russell, 2005] N. Oza and S. Russell, "Online bagging and boosting", *in Proceedings of the International Conference on Systems, Man and Cybernetics*, pages 2340-2345, 2005.

[Yang *et al.*, 2009] L. Yang, R. Jin, J Ye. Online learning by ellipsoid method. *in Proceedings of the 26th International Conference on Machine Learning*, Montreal, Canada, 2009

[Johnson and Lindenstrauss, 1984] W. Johnson, J. Lindenstrauss. Extensions of Lipshitz mapping into Hilbert space. *in Proceeding of the Conference in modern analysis and probability*. 26, pages 189-206, American Mathematical Society, 1984.

[Bingham and Mannila, 2001] E. Bingham, H. Mannila. Random projection in dimensionality reduction: applications to image and text data. *in Proceeding of the 7th International Conferene on Knowledge Discovery and Data Mining (ACM SIGKDD)*, pages 245-250, 2001.

[Fern and Brodley, 2003] X.Z. Fern, C.E. Brodley, Random Projection for High Dimensional Data Clustering: A Cluster Ensemble Approach, *in Proceedings of the 20th International Conference on Machine Learning (ICML)*, pages 186-193, 2003.

[Avogadri and Valentini, 2009] R. Avogadri, G. Valentini. Fuzzy ensemble clustering based on random projections for DNA microarray data analysis. *Artificial Intelligence in Medicine*, 45: 173-183, 2009.

[Domingo, Pazzani, 1996] P. Domingos, M. Pazzani, Beyond independence: Conditions for the optimality of the simple Bayesian classifier, In Proceedings of the Thirteenth International Conference on Machine Learning, pp. 105–112, 1996.

[Nguyen *et al.*, 2014] T.T. Nguyen, A.W.-C. Liew, M.T. Tran, X.C. Pham, M.P. Nguyen, A novel genetic algorithm approach for simultaneous feature and classifier selection in multi classifier system, *in Proceeding of the IEEE Congress on Evolutionary Computation* (CEC), pages 1698-1705, 2014.

[Dietterich, 2000] T. Dietterich, Ensemble methods in machine learning, in the first International Workshop on Multiple Classifier Systems, Springer, pp. 1-15, 2000.

[Nguyen *et al.*, 2016b] T.T. Nguyen, T.T.T. Nguyen, X.C. Pham, A.W-C. Liew, A Novel Combining Classifier Method based on Variational Inference, *Pattern Recognition*, 49: 198-212, 2016.

[Balakrishnan, Nevzorov, 2003] N. Balakrishnan, V.B. Nevzorov, A Primer on Statistical Distributions, Wiley &Sons Press, 2003

[Kittler *et al.*, 1998] J. Kittler, M. Hatef, R.P.W. Duin, J. Matas, On Combining Classifiers, *IEEE Transactions on Pattern Analysis and Machine Intelligence*, 20(3): 226-239, 1998.

[UCI] UCI Machine Learning Repository: Data Sets, http://archive.ics.uci.edu/ml/datasets.html

[Hoi *et al.*, 2014] S.C.H. Hoi, J. Wang, and P. Zhao. LIBOL: A Library for Online Learning Algorithms. *Journal of Machine Learning Research.* 15: 495-499, 2014.

[Bifet *et al.*, 2010] A. Bifet, G. Holmes, B. Pfahringer, P. Kranen, H. Kremer, T. Jansen, T. Seidl. MOA: Massive Online Analysis, a Framework for Stream Classification and Clustering. *in Journal of Machine Learning Research Workshop and Conference Proceedings*, Vol. 11: Workshop on Applications of Pattern Analysis, 2010.

[Sokolova and Lapalme, 2009] M. Sokolova, G. Lapalme, A systematic analysis of performance measures for classification tasks, *Information Processing and Management*.45: 427-437, 2009.

[Demsar, 2006] J. Demsar, Statistical comparisons of classifiers over multiple datasets, *Journal of Machine Learning Research.* 7:1-30, 2006).

[Webb *et al.*, 2005] G. I. Webb, J. R. Boughton, Z. Wang. Not So Naïve Bayes: Aggregating One-Dependence Estimators. *Machine Learning*. 58: 5-24, 2005.